\documentclass[conference]{IEEEtran} 
\IEEEoverridecommandlockouts
\usepackage{cite}
\usepackage{lmodern}
\usepackage{graphicx}
\usepackage{times}
\usepackage{ragged2e}
\usepackage{amsmath,amssymb, amsfonts}
\usepackage{tabularx}
\usepackage{hyperref}
\usepackage{cleveref}
\usepackage[ruled,vlined]{algorithm2e}
\usepackage{siunitx}
\usepackage{dblfloatfix}
\usepackage{authblk}
\usepackage[table]{xcolor}
\usepackage[T1]{fontenc}
\usepackage{siunitx}
\usepackage{booktabs}
 
\begin{document}
\noindent

\title{\LARGE \bf
Robust LSTM-based Vehicle Velocity Observer for Regular and Near-limits Applications
}

\author{
Agapius Bou Ghosn$^{1}$,
Marcus Nolte$^{2}$,
Philip Polack$^{1}$,
Arnaud de La Fortelle$^{1,3}$,
and Markus Maurer$^{2}$
\thanks{$^{1}$ Center for Robotics, Mines Paris, PSL University, 75006 Paris, France {\tt [agapius.bou\textunderscore ghosn, philip.polack, arnaud.de\textunderscore la\textunderscore fortelle]@minesparis.psl.eu}}
\thanks{$^{2}$ Institute for Control Engineering, TU Braunschweig, 38106 Braunschweig, Germany {\tt [nolte, maurer]@ifr.ing.tu-bs.de}}
\thanks{$^{3}$ Heex Technologies, Paris, France}
}

\maketitle

\thispagestyle{empty}
\pagestyle{empty}

\begin{abstract}
Accurate velocity estimation is key to vehicle control. While the literature describes how model-based and learning-based observers are able to estimate a vehicle's velocity in normal driving conditions, the challenge remains to estimate the velocity in near-limits maneuvers while using only conventional in-car sensors. In this paper, we introduce a novel neural network architecture based on Long Short-Term Memory (LSTM) networks to accurately estimate the vehicle's velocity in different driving conditions, including maneuvers at the limits of handling. The approach has been tested on real vehicle data and it provides more accurate estimations than state-of-the-art model-based and learning-based methods, for both regular and near-limits driving scenarios. Our approach is robust since the performance of the state-of-the-art observers deteriorates with higher dynamics, while our method adapts to different maneuvers, providing accurate estimations even at the vehicle's limits of handling.    
\end{abstract}

\section{Introduction}
The knowledge of vehicle dynamics states is crucial for successful autonomous driving applications. It is the basis for proper planning and control. While many of the state variables of a vehicle are measurable through in-car conventional sensors, the knowledge of the non-measurable state variables, and the filtering of the noisy measured states necessitates the use of estimation tools: State observers.   

Classical state observers rely on a model to describe the evolution of system states and take into account the available inputs and measurements to output the state estimations. The quality of the estimations is hence highly influenced by the model describing the system: It should be as close as possible to the actual system. On the other hand, data-driven approaches are independent of explicit physical models, relying on data collected from the vehicle to estimate the needed quantities.    

The longitudinal and lateral velocities of a vehicle are examples of non-measurable state variables describing a vehicle's dynamic behavior. 
Knowledge about these states, along with the vehicle's yaw rate, is key to effective vehicle control. 

The observers presented in the literature are able to estimate these quantities within a certain level of accuracy for normal driving scenarios. The challenge remains in developing an observer able to perform accurately in both low acceleration and high acceleration maneuvers (cf. Section \ref{relatedWorks.sec}). 

\begin{figure}
    \centering
    \includegraphics[width=.8\columnwidth]{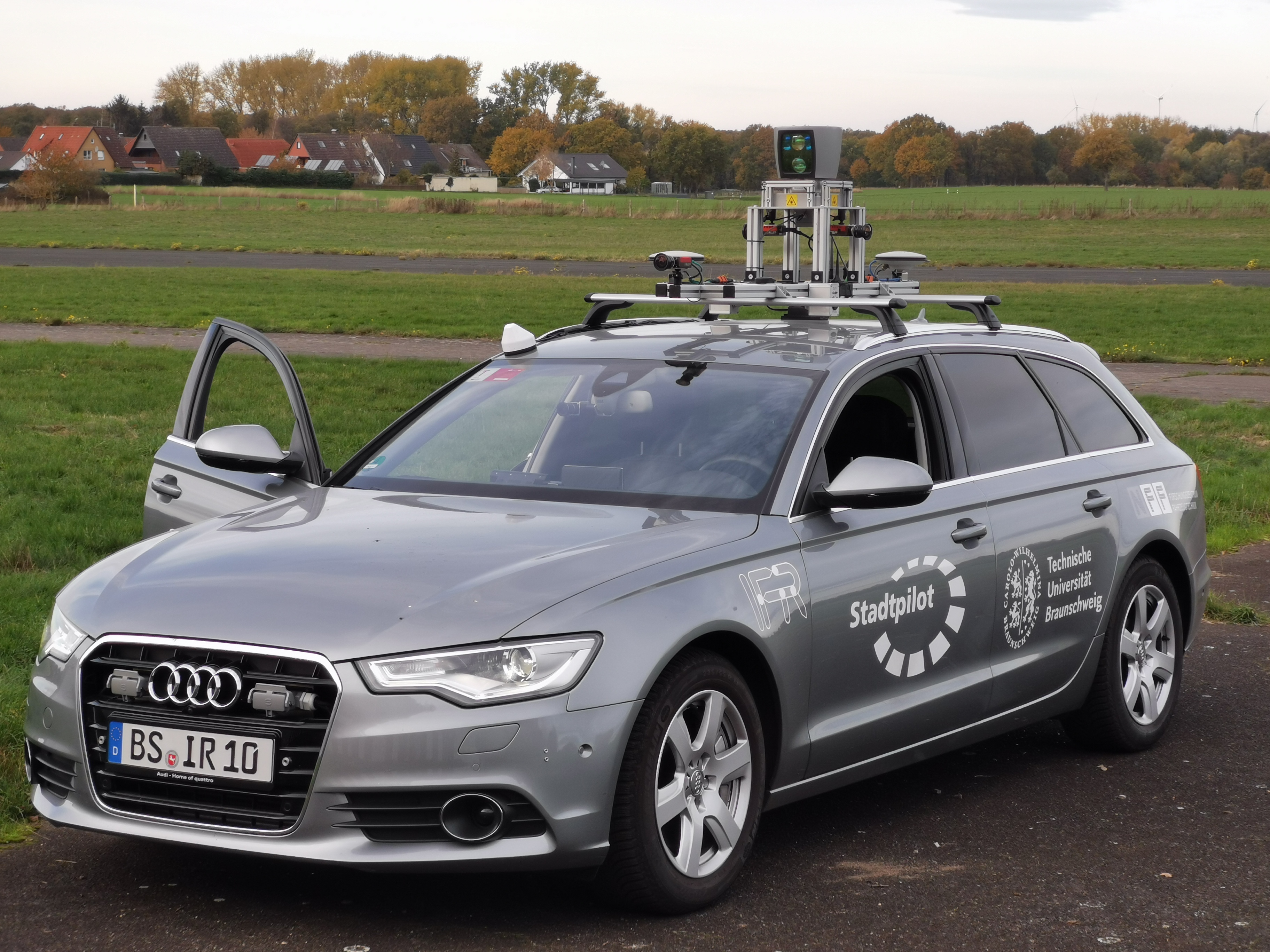}
    \caption{Stadtpilot vehicle used for testing the presented approach.}
    \label{vehicles.fig}
\end{figure}

In this paper, we propose an observer architecture based on LSTM networks.
This architecture is able to perform longitudinal velocity, lateral velocity and yaw rate estimations in normal and dynamically challenging driving scenarios (up to $a_y=0.8g$) using only conventional in-car sensors. The technique is demonstrated on the Stadtpilot vehicle of TU Braunschweig shown in Fig.~\ref{vehicles.fig} and tested on thoroughly performed maneuvers. The performance is compared to state of the art observers presented in the next section.

In summary, this paper presents the following contributions:
\begin{itemize}
    \item We present an LSTM-based observer to accurately estimate the longitudinal velocity, lateral velocity and yaw rate of a vehicle in normal and near-limits driving conditions using only in-car sensor measurements.
    \item We validate the performance of the proposed learned observer on a real vehicle while comparing to high accuracy ground truth sensors.
    \item We demonstrate that the presented learned observer outperforms state of the art model-based and learned-based observers.   
\end{itemize}

In the following, we present the state of the art observers in Section \ref{relatedWorks.sec}. We describe the system setup in Section \ref{systemSetup.sec} as well as the data set collection for training and testing the observers in Section \ref{dataset.sec}. Section \ref{approach.sec} presents the proposed architecture and its training, while  Section \ref{results.sec} shows the results including the comparison to state of the art methods. The paper is concluded in Section \ref{conclusion.sec}.

\section{Related Work}\label{relatedWorks.sec}
As we create a learning based observer to estimate the vehicle's velocities and yaw rate, the state of the art observers should be explored. Our review is split into model-based and learning-based (and hybrid) observers. We present model-based state of the art observers in Section \ref{mb_obs.ssec} and learning-based observers in Section \ref{lb_obs.ssec}. 

\subsection{Model-based observers}\label{mb_obs.ssec}
Several classical approaches, with different observing techniques and vehicle models were used in the literature to estimate the vehicle's velocities. The Luenberger linear observer has been employed in applications like \cite{cherouat_vehicle_2005}, \cite{kiencke_observation_1997} to estimate the vehicle velocity, side slip angle, and yaw rate using a dynamic bicycle model with a linear tire model. The state of the vehicle is accurately estimated in normal driving scenarios while inaccuracies occur for more dynamic maneuvers as soon as non-linear dynamics start dominating the models' behavior.  Non-linear observers with more complex vehicle and tire models have e.g. been employed in \cite{zhao_design_2011}, \cite{imsland_vehicle_2006}; In these works, the model used is able to describe the vehicle's maneuvers in a more accurate way for higher dynamics leading to more accurate estimations. However, approaches based on the Luenberger observer have the inherent disadvantage to heavily rely on accurate measurements and models, as they do not account for measurement- or process noise.

The Kalman filter is used in many estimation applications. While there are non-linear extensions, in its original form, it is a linear 
optimal filter. The Kalman filter algorithm involves a prediction step where the current state and covariances of the system are calculated based on the previously estimated state, the current inputs and covariances and an update step where the current measurements are used to correct the calculated state and covariances.
Many, also fairly recent, works implemented linearized or non-linear variations of the Kalman Filter to estimate the state of the vehicle; the Extended Kalman Filter (EKF) for example has been used along with kinematic or dynamic bicycle models (\cite{dickmanns1992}, \cite{kim_vehicle_2018}, \cite{van_aalst_adaptive_2018}, \cite{reina_vehicle_2019}) to estimate the vehicle's velocities, yaw rate, side-slip angle and tire forces with different considerations to the tire model. Also, it was used with a four wheel dynamic model \cite{katriniok_adaptive_2016} and a Pacejka tire model to estimate the vehicle's velocities and yaw rate.

The presented model-based observers are able to estimate the state of the vehicle in a specific operational domain where the assumptions of their used models are still valid; the challenge is to estimate the state of the vehicle beyond these assumptions. 

The approaches presented in \cite{van_aalst_adaptive_2018} and \cite{katriniok_adaptive_2016} show accurate estimations on the scenarios they were tested on; they will be compared to the proposed approach.

\subsection{Learning-based observers}\label{lb_obs.ssec}
Machine learning based observers have been proposed in the recent years to estimate the vehicle's state based on measurements data, e.g. to overcome limitations due to insufficient knowledge about model parameters. When it comes to the vehicle's velocities, the literature includes hybrid approaches that combine model-based and learned observers as in \cite{escoriza_data-driven_2021} where a KalmanNet \cite{revach_kalmannet_2021} architecture is used for velocity estimation. This architecture implements a Kalman filter and replaces the gain calculation by a recurrent neural network prediction. Similarly, \cite{noauthor_vehicle_2021} use a neural network along a sliding mode observer and a Kalman filter for velocity estimation. Other works use Long Short-Term Memory (LSTM) or Gated Recurrent Units (GRU) networks, as in \cite{liu_vehicle_2021}, \cite{zhang_reliable_2021} and \cite{srinivasan_end--end_2020} where longitudinal and lateral velocities are predicted from multiple sensor measurements, or as in \cite{wang_deepspeedometer_2017} where the speed of the vehicle is predicted based on IMU measurements. The presented learned observers show accurate observation quality.
However, in contrast to our proposed solution, they necessitate additional measurements in many cases (e.g. GPS measurements \cite{noauthor_vehicle_2021}, throttle position sensors \cite{liu_vehicle_2021}, longitudinal velocity measurements \cite{zhang_reliable_2021}). 

The approaches presented in \cite{escoriza_data-driven_2021} and \cite{srinivasan_end--end_2020} show accurate estimations for multiple driving scenarios and thus will be used for comparison  when evaluating our method.     

\section{System Setup}\label{systemSetup.sec}
As we present a learning based observer for autonomous vehicle applications, we made sure that the approach is applied on a real vehicle. For that, we use the Stadtpilot vehicle (AUDI A6 Avant C7) shown in Fig.~\ref{vehicles.fig} for data collection. The characteristics of the vehicle are shown in \Cref{vehicleChs.tab}.

\begin{table}
\centering
%  \resizebox{\columnwidth}{!}{
    \caption{Parameters of the vehicle used for data collection (CoG: center of gravity)}
    \begin{tabular}{cp{.5\columnwidth}c}
      \toprule
      \textbf{Parameter} & \textbf{Description} & \textbf{Value} \\
      \midrule
      $M$ & Mass of the vehicle & \SI{1578}{\kilo\gram}\\
      $l_\mathrm{f}$ & length from CoG to the front axle & \SI{1.134}{\meter} \\
      $l_\mathrm{r}$ & length from CoG to the rear axle  & \SI{1.578}{\meter} \\
      $b$ & Track width & \SI{1.513}{\meter} \\
      $I_z$ & Moment of inertia around the $z$-axis & \SI{2924}{\kilo\gram\square\meter}\\
      \bottomrule
    \end{tabular}
%}
\label{vehicleChs.tab}
\end{table}

\begin{figure}
    \centering
    \includegraphics[width=\columnwidth]{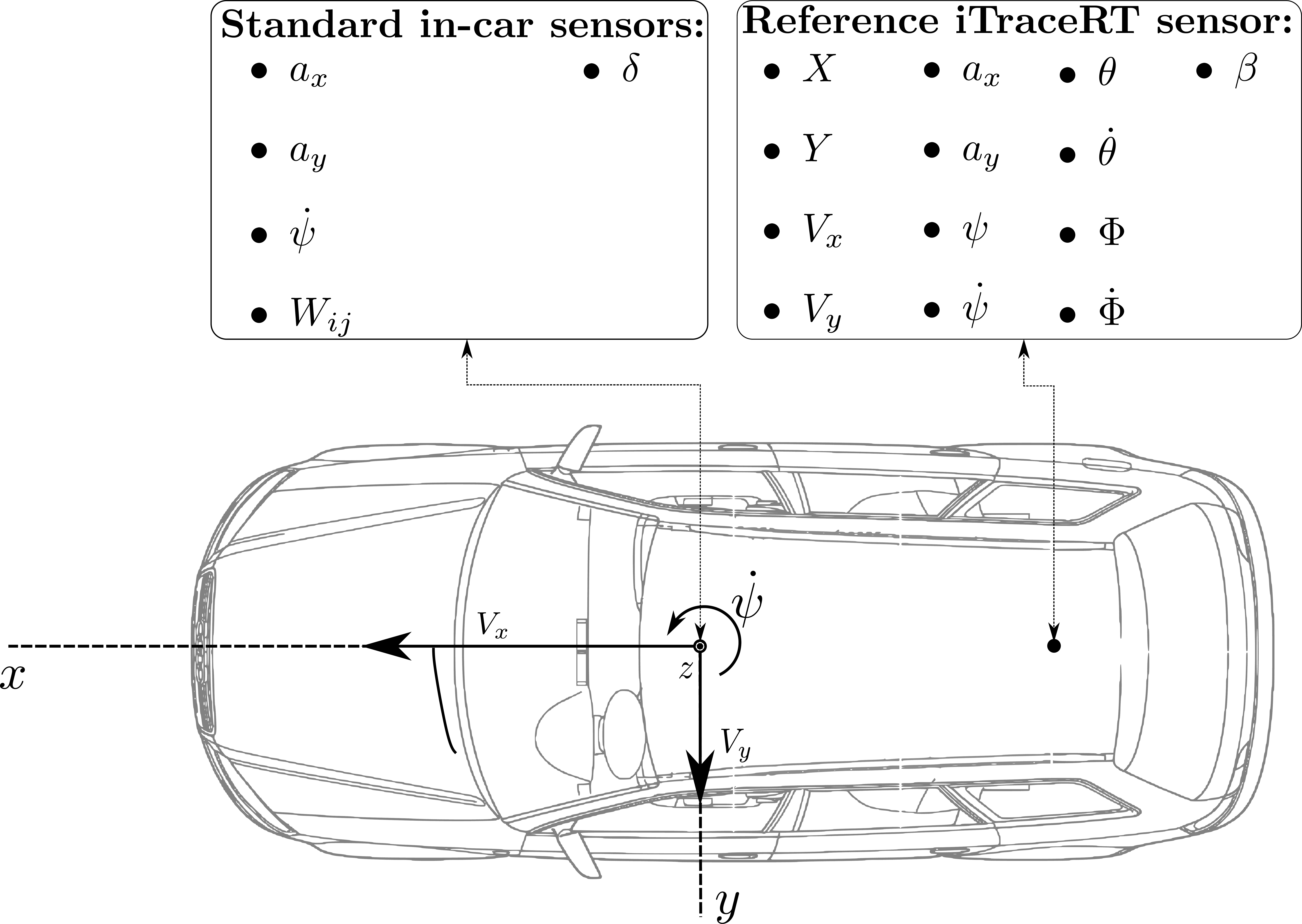}
    \caption{Top view of the used vehicle showing the available in-car and reference sensor measurements. The longitudinal $V_x$ and lateral $V_y$ velocities and the yaw rate $\dot\psi$ shown at the center of gravity of the vehicle will be estimated. The variables are described in Section \ref{systemSetup.sec}.}
    \label{topView.fig}
\end{figure}

The vehicle is equipped with the Audi Sensor Array (SARA) which provides longitudinal and lateral accelerations ($a_x$, $a_y$), the yaw rate ($\dot\psi$), the wheel speeds ($W_{ij}$) and the steering angle ($\delta$). The reference dual-antenna INS/GNSS \href{https://www.imar-navigation.de/en/products/by-product-names/item/itracert-f200-itracert-f400-itracert-mvt}{iTraceRT F400} sensor is mounted to the vehicle to give accurate measurements for the evolution of its state; it provides the position (Easting -- $X$ , Northing -- $Y$) in UTM-coordinates, the longitudinal and lateral velocities ($V_x$, $V_y$), the longitudinal and lateral accelerations ($a_x$, $a_y$), the yaw ($\psi$), pitch ($\theta$), and roll ($\Phi$) angles and rates ($\dot\psi$, $\dot\theta$, $\dot\Phi$), and the side-slip angle ($\beta$). The top view of the vehicle is shown in Fig.~\ref{topView.fig} showing the available sensors and the variables of interest at the center of gravity: the longitudinal and lateral velocities and the yaw rate. The in-car sensors provide measurements at a \SI{50}{Hz} frequency while the reference sensor provides measurements at \SI{100}{Hz}; The reference sensor measurements will be down sampled to \SI{50}{Hz} for synchronization purposes. 
The built learning-based observer will take its input data from the in-car sensors and will compare its outputs to the reference sensor measurements which are considered as the ground truth. 
The defined system will be used for data collection next. 

\section{Data Set}\label{dataset.sec}
The previously defined system is used for data collection. The recorded data set contains two types of maneuvers: \begin{enumerate}
    \item Low acceleration maneuvers effected in the city of Braunschweig, Germany in normal weather conditions during which multiple maneuvers were considered: U-Turns, lane changes, in addition to normal driving through out the city. This part consists of 770,000 samples equivalent to 4.3 hours of driving.
    \item High acceleration maneuvers executed on a special test track near Peine, Germany consisting of 300,000 samples equivalent to 1.6 hours of driving.
\end{enumerate}
The total adds up to about 1 million samples.

\begin{figure}
    \centering
    \includegraphics[width=\columnwidth]{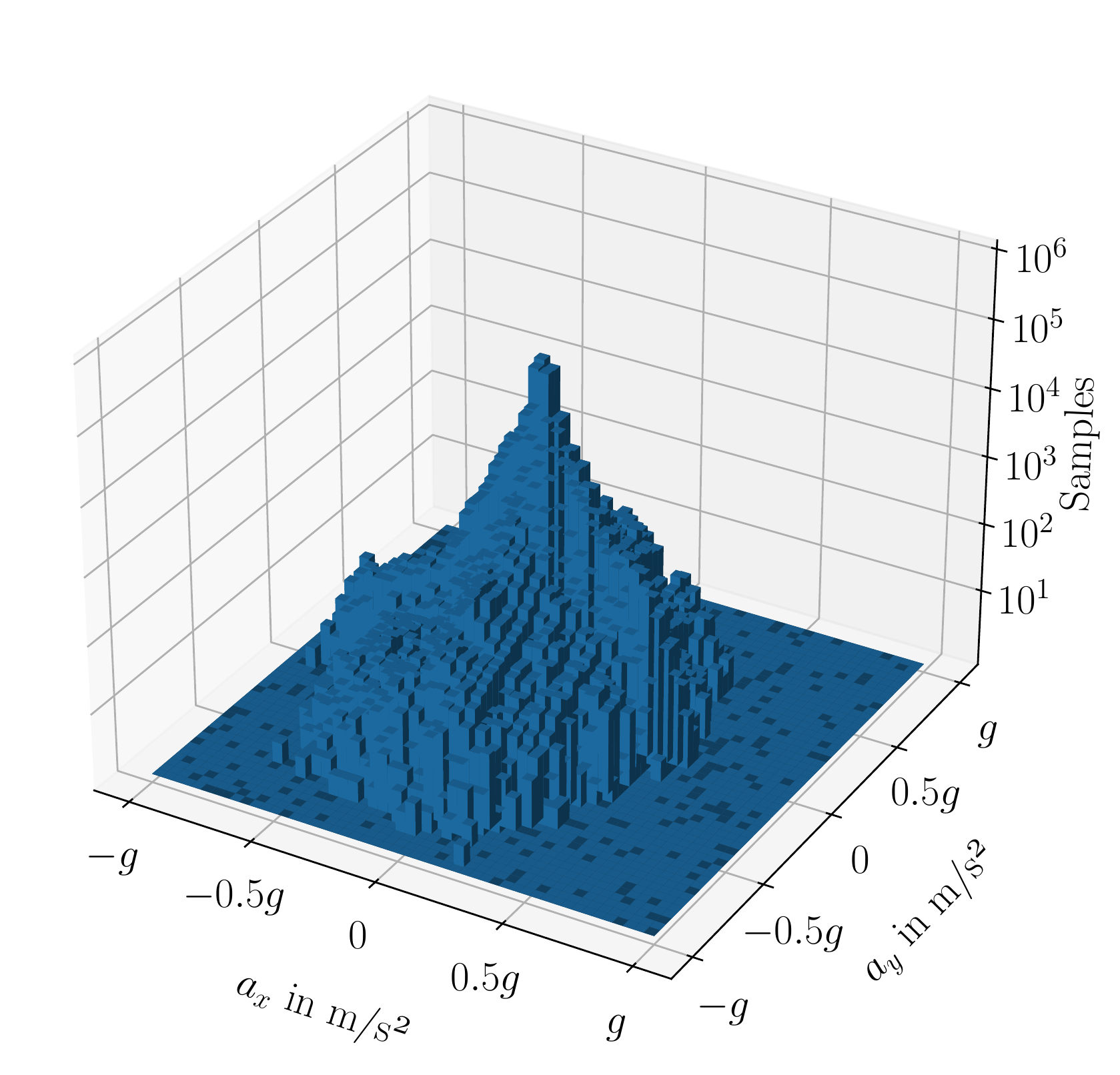}
    \caption{Distribution of the collected data on the friction circle (number of samples shown on a logarithmic axis): Most of the samples are in a low acceleration range, but samples are present at high accelerations. $g$ is the gravitational acceleration expressed in $\SI{}{m/s^2}$.}
    \label{frictionCircle.fig}
\end{figure}

To inspect the characteristics of the collected dataset, we plot the resulting friction circle in Fig.~\ref{frictionCircle.fig}. The plot shows that the distribution is more populated in the low acceleration regions which is rather logical: these values correspond to the normal city driving in addition to the transitions between high acceleration maneuvers; high acceleration data points (up to $a_y=1g$) are present in the data set, as well: these correspond to harsh driving maneuvers executed on the test track.  

Having collected the needed data, a 60\%-20\%-20\% split is applied resulting in training, validation and testing data sets respectively; the split is carefully performed to include both low acceleration and high acceleration data in all sets.

The observing approach is defined next.

\section{Proposed Approach}\label{approach.sec}
The aim of the developed observer is to estimate the current longitudinal and lateral velocities and the yaw rate of the vehicle given the in-car sensor measurements. Having the ground truth iTraceRT sensor measurements, the target values are available to train a learning-based observer.

Recurrent neural networks are chosen as the basis for our learned observer as they are able to accurately learn dynamic systems through their dependencies on previous time steps \cite{graber_hybrid_2019}. LSTMs were developed to deal with the exploding/vanishing gradients problem associated with traditional recurrent neural networks \cite{hochreiter_long_1997}; they are implemented in our proposed architecture. 

We propose an LSTM-based observer that takes two inputs:

The first input is the in-car sensor measurements for the 50 previous time steps. Using multiple previous measurements provides the neural network with additional information to adapt to the actual vehicle dynamics. The choice of the number of time steps is based on the approach shown in \cite{bou_ghosn_learning-based_2022}. Note that we choose one of the wheel speeds (rear right) to be used from the in-car measurements: this ensures a fair comparison with the state of the art methods that make this choice. 

The second input includes the previous vehicle velocities and yaw rate giving the observer knowledge about the state of the vehicle in the previous time step. 
Note that we use two different sources for the input velocities and yaw rate during training and testing, as will be discussed below.
%It should be noted that the second input will be fed from the predictions of the observer during testing; in other words, the second input consists of the previous vehicle state (taken from the ground truth data) when training and it consists of the previous observer estimation when testing.

The output of the observer is the current vehicle's longitudinal and lateral velocities and yaw rate.

\begin{figure}
    \centering
    \includegraphics[width=\columnwidth]{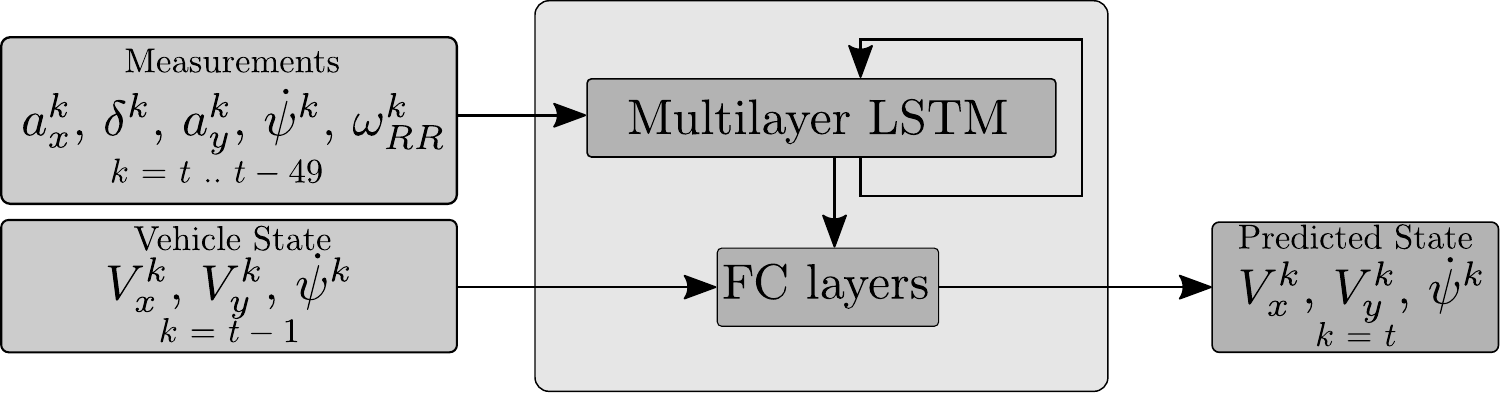}
    \caption{The LSTM-based observer during training. The network takes two inputs: the in-car sensor measurements and the vehicle velocities and yaw rate at the previous time step. The outputs are the estimated vehicle velocities and yaw rate at the current time step.}
    \label{lstm_training.fig}
\end{figure}

We specify two operating modes of our architecture:

The training mode shown in Fig.~\ref{lstm_training.fig}: In this mode, the vehicle state data from the previous time step fed to the network is the ground truth data to which Gaussian noise is added; the added noise makes the network immune to drifting as it will be trained to predict the current state even if the previous state includes some noise (Which could be the case in the testing mode). The standard deviations for the added noise are $\SI{0.03}{m/s}$ for both longitudinal and lateral velocities and $\SI{0.003}{rad/s}$ for the yaw rate.

\begin{figure}
    \centering
    \includegraphics[width=\columnwidth]{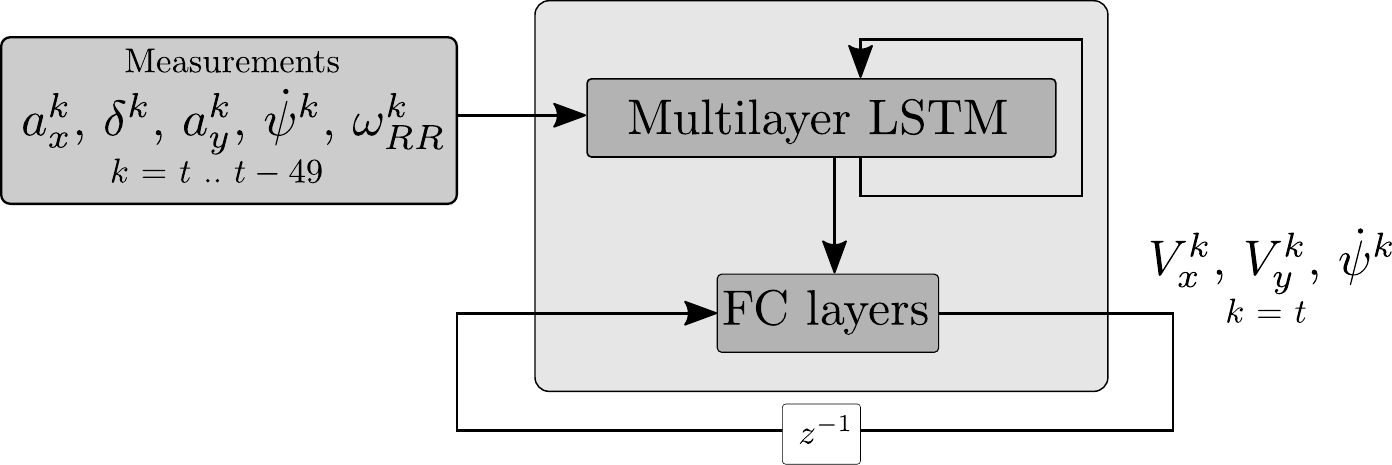}
    \caption{The LSTM-based observer during testing. The first input to the network is still the measurements from the in-car sensors while the second input are the predictions of the observer at the previous time step.}
    \label{lstm_testing.fig}
\end{figure}

The testing mode shown in Fig.~\ref{lstm_testing.fig}: In this mode, the data for the previous time steps is the previous predictions of the observer, which will create a closed loop. At the beginning of its operation, an initial state should be provided to the observer.

The architecture shown in Fig.~\ref{lstm_testing.fig} shows the operating mode of our observer. The architecture includes a set of LSTM layers to which the measurements of the in-car sensors are fed. The LSTM part of the network is made of four LSTM layers including $32$, $64$, $64$ and $128$ neurons respectively. Fully connected layers follow, to which the output of the last LSTM layer and the previous state are fed; the three fully connected layers have $64$, $128$ and $64$ neurons respectively. The output of the network is the current longitudinal velocity, lateral velocity and yaw rate estimations. The inputs and outputs of the network are scaled between 0 and 1. The fully connected layers have sigmoid activation functions. An $L2$ loss function is used when training. The network is implemented using PyTorch and is trained on an Nvidia Geforce GTX 1650 Ti for 50 epochs. After training the network, we proceed to testing it and comparing its performance to state of the art approaches.

\section{Results}\label{results.sec}
The developed approach in addition to the state of the art approaches presented in Section \ref{relatedWorks.sec} will be tested on the testing data set presented in Section \ref{dataset.sec}. 

In the following, the mean absolute error (MAE) metric will be used. All of the observers will be first evaluated on the whole testing set; then, separate scenarios will be considered: a normal driving scenario and a near-limits scenario. The performance difference between the observers will be highlighted. Note that the notion of ranking stated in the following paragraphs consists of ranking first the observer with the lowest errors and ranking last the observer with the highest errors. 

\subsection{Overall performance}
To evaluate the performance of the different observers,  we implemented the respective approaches and applied them to our testing dataset. The mean absolute errors have been calculated for each of the estimated variables. \Cref{res1.tab} shows the errors for the different observers. The errors show that our approach is able to estimate the three variables with the lowest errors. We remark that the KalmanNet approach \cite{escoriza_data-driven_2021} ranks second and the EKF based on the dynamic bicycle model \cite{van_aalst_adaptive_2018} ranks last for the longitudinal velocity estimation; the GRU-based approach \cite{srinivasan_end--end_2020} ranks second and the KalmanNet approach \cite{escoriza_data-driven_2021} ranks last for the lateral velocity estimation; the EKF based on the dynamic bicycle model \cite{van_aalst_adaptive_2018} ranks second and the GRU-based approach \cite{srinivasan_end--end_2020} ranks last for the yaw rate estimation. To further inspect the performance of the different observers, we look into specific maneuvers from the testing set. Next, the performance of the observers for a sequence of normal driving maneuvers is explored.

\begin{table}
\centering
\setlength\tabcolsep{5 pt}
    \caption{Mean absolute error (MAE) for the different observers calculated for the whole testing set. The errors of the proposed approach are the lowest among the state of the art observers. \textbf{Legend:} DBM: EKF based on the dynamic bicycle model; 4WM: EKF based on the four wheel vehicle model; KN: KalmanNet based observer; GRU: End-to-end GRU-based observer. \textbf{Color code:} Green cells indicate the lowest errors. Orange cells indicate the next to lowest errors. Red cells indicate the largest errors.}
 %\resizebox{\columnwidth}{!}{
    \begin{tabular}{|c|c|c|c|c|c|}
      \hline
      State & DBM \cite{van_aalst_adaptive_2018} & 4WM \cite{katriniok_adaptive_2016}  & KN \cite{escoriza_data-driven_2021}  & GRU \cite{srinivasan_end--end_2020}  & Ours\\
      \hline
      $V_x$ \SI{}{(m/s)} &\cellcolor{red} 0.2 & 0.072 & \cellcolor{orange} 0.045 & 0.054 & \cellcolor{green!70}  \textbf{0.040} \\
      \hline
      $V_y$ \SI{}{(m/s)} & 0.052 & 0.038 &\cellcolor{red} 0.095 & \cellcolor{orange} 0.023 & \cellcolor{green!70} \textbf{0.021} \\
      \hline
      $\dot\psi$ \SI{}{(mrad/s)} & \cellcolor{orange} 4.51 & 4.52 & 4.54 & \cellcolor{red} 4.68 & \cellcolor{green!70} \textbf{2.94} \\
      \hline
    \end{tabular}
%}
\label{res1.tab}
\end{table}

\subsection{Normal driving maneuvers}
\begin{figure}
    \centering
    \includegraphics[trim={0.7em 0 0 0}, clip=true, width=\columnwidth]{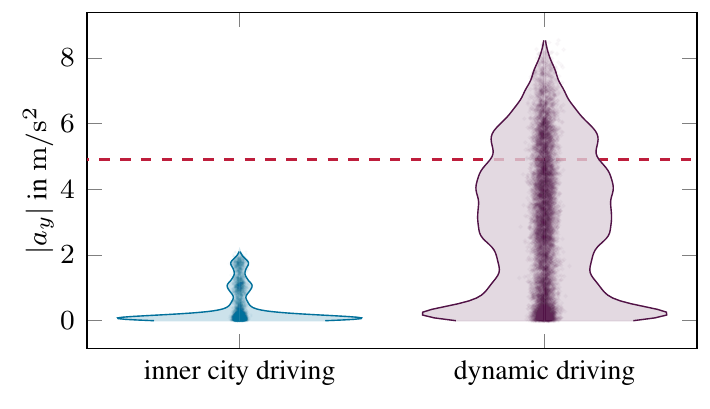}
    \caption{Violin plot showing the distribution of lateral accelerations in the datasets used for testing. Compared to the inner city driving test set, the dynamic driving test set contains samples beyond the lateral accelerations of the inner city driving test set \emph{and} the $0.5g$-limit discussed by \cite{polack_kinematic_2017}.}
    \label{normalTraj.fig}
\end{figure}

After assessing the overall performance, we consider a sequence of non-dynamic driving maneuvers to compare the different observers. The distribution of the absolute lateral accelerations during the maneuver sequence are plotted in Fig.~\ref{normalTraj.fig}. The dotted line represents $a_y=0.5g$ beyond which many vehicle models become invalid due to the violation of linearity assumptions (mainly in the tire models) (\cite{kiencke_observation_1997}, \cite{polack_kinematic_2017}). The acceleration plot shows that the maximum acceleration is $a_y^{\text{max}} = \SI{2.10}{m/s^2}$  indicating low dynamic driving. The MAE of the different observers for the considered maneuver are shown in \Cref{res2.tab}. The proposed approach is able to perform best among the considered state of the art observers for the three variables. 

It can be seen that for the longitudinal velocity estimation, the KalmanNet based approach \cite{escoriza_data-driven_2021} ranks second with an estimation error that is close to our approach; the EKF based on the dynamic bicycle model \cite{van_aalst_adaptive_2018} shows the worst performance with an error 3 times higher than the proposed approach. For the lateral velocity estimation, the GRU-based approach \cite{srinivasan_end--end_2020} ranks second; the KalmanNet based approach \cite{escoriza_data-driven_2021} performs the worst with an error 5 times higher than the proposed approach. For the yaw rate estimation, the errors of the presented state of the art approaches are similar except for the GRU-based \cite{srinivasan_end--end_2020} approach that performs the worst.

\begin{figure}
    \centering
    \includegraphics[trim={0.5em 0 0 0}, clip=true, width=\columnwidth]{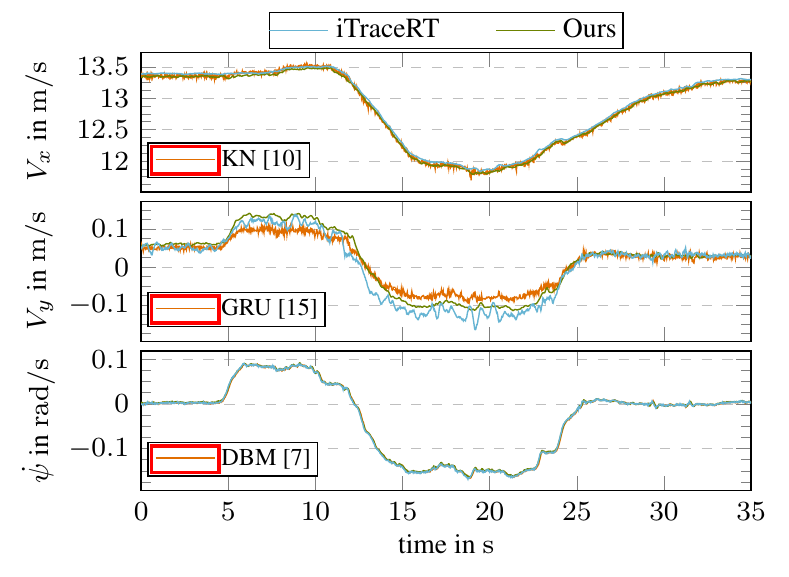}
    \caption{Comparison between the proposed approach, the next best performing approach and the reference for normal driving for each of the variables. The performance is close for $V_x$ and $\dot\psi$ but a clear advantage for the proposed method can be seen for $V_y$. Note that the proposed method is able to provide the most accurate results for the three variables of interest while this is not the case for the other observers (cf. \Cref{res2.tab}).} \label{perf_lM1.fig}   
\end{figure}

To further inspect the presented results, we plot the estimations by our proposed approach and the next best performing approach of each of the variables. We plot them in a subset of the maneuver sequence summarized in Fig.~\ref{normalTraj.fig}. This subset includes the highest lateral accelerations in the dataset. All approaches are compared to the iTraceRT sensor values for reference. Higher accelerations are associated with more dynamic maneuvers and thus more challenging estimations. The plot is shown in Fig.~\ref{perf_lM1.fig}. It shows a close performance between our approach and the considered state of the art approaches for both the longitudinal velocity and yaw rate estimations. This is not the case for the lateral velocity estimation; the lateral velocity plot shows a clear advantage for our approach which is able to closely follow the reference iTraceRT values. Note that none of the considered state of the art approaches is able to achieve low errors for the three variables simultaneously. 
Next, a near-limits maneuver is considered.

\begin{table}
\centering
\setlength\tabcolsep{5 pt}
    \caption{Mean absolute error for the different observers calculated during normal driving. The errors of the proposed approach are the lowest among the state of the art observers. Cf. \Cref{res1.tab} for legend and color codes.}
    \begin{tabular}{|c|c|c|c|c|c|}
      \hline
      State & DBM \cite{van_aalst_adaptive_2018} & 4WM \cite{katriniok_adaptive_2016}  & KN \cite{escoriza_data-driven_2021}  & GRU \cite{srinivasan_end--end_2020}  & Ours\\
      \hline
      $V_x$ \SI{}{(m/s)} & \cellcolor{red} 0.12 & 0.084 & \cellcolor{orange} 0.041 & 0.059 & \cellcolor{green!70} \textbf{0.039} \\
      \hline
      $V_y$ \SI{}{(m/s)} &  0.049 &  0.038 & \cellcolor{red} 0.055 & \cellcolor{orange} 0.014 & \cellcolor{green!70} \textbf{0.011} \\
      \hline
      $\dot\psi$ \SI{}{(mrad/s)} & \cellcolor{orange} 2.15 &  2.48 & 2.52 & \cellcolor{red} 4.02 & \cellcolor{green!70} \textbf{2.11} \\
      \hline
    \end{tabular}
    %}
\label{res2.tab}
\end{table}

\subsection{Near-limits maneuvers}
Having inspected the behavior of the observers for a normal driving scenario, we consider a sequence of near-limits maneuvers. The maneuvers were performed on a dedicated test track to reach high accelerations. The distribution of absolute lateral accelerations throughout the maneuver sequence is seen in Fig.~\ref{normalTraj.fig} showing harsh behavior with values reaching $a_y = 0.8g$. 

The MAE of the different observers are presented in \Cref{res4.tab}; It can be seen that for all of the observers the errors increase for this maneuver comparing with previous maneuvers, this is due to the harshness of the maneuver. The proposed approach delivers the best performance among the considered state of the art approaches.

It can be seen that for both the longitudinal and lateral velocity estimations, the GRU-based approach \cite{srinivasan_end--end_2020} ranks second while the EKF based on the dynamic bicycle model \cite{van_aalst_adaptive_2018} gives the worst performance. For the yaw rate estimation, the EKF based on the dynamic bicycle model \cite{van_aalst_adaptive_2018} ranks second while the KalmanNet based approach \cite{escoriza_data-driven_2021} performs worst with an error that is almost double the error of the proposed approach. 

It is remarked that the GRU-based observer \cite{srinivasan_end--end_2020} ranks second in both maneuvers for the lateral velocity estimation and that the EKF based on the dynamic bicycle model \cite{van_aalst_adaptive_2018} ranks second in both maneuvers for the yaw rate estimation. For the longitudinal velocity estimation the KalmanNet observer ranks third in the harsh maneuver while it ranks second in the normal driving maneuver; the GRU-based observer ranks second in the harsh maneuver while it ranks third in the normal driving maneuver. The proposed approach ranks first for all the variables in both maneuvers. 
 
\begin{figure}
    \centering
    \includegraphics[trim={0.5em 0 0 0}, clip=true, width=\columnwidth]{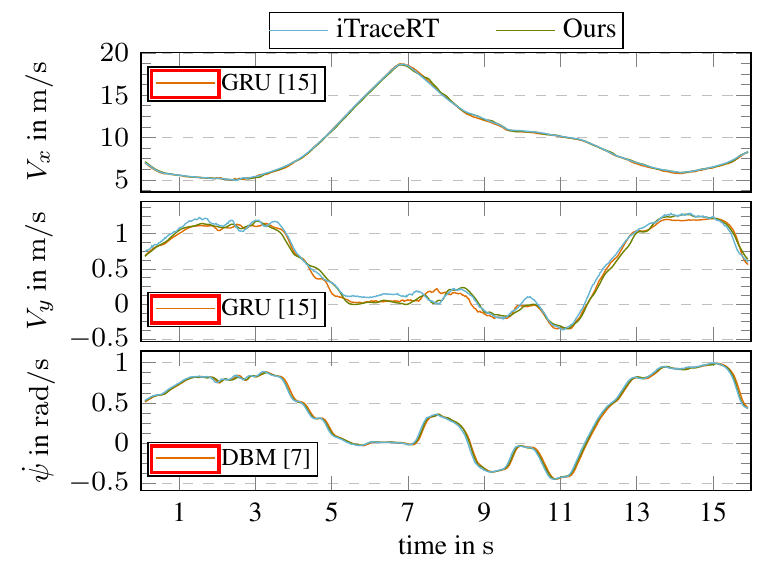}
    \caption{Comparison between the proposed approach, the next best performing approach and the reference for near-limits driving for each of the variables. The performance is close for $V_x$ but a clear advantage for the proposed method can be seen for $V_y$ and $\dot\psi$. Note that the proposed approach is able to provide the most accurate results for all three state variables of interest, while this is not the case for the other observers (cf. \Cref{res4.tab}). }
    \label{perf_hM2.fig}
\end{figure}

To further inspect the presented results, we again plot the estimations by the proposed approach and the next best performing approach of each of the variables for the highest lateral acceleration point in the maneuver sequence. The iTraceRT sensor values are again used as a reference. The plot is shown in Fig.~\ref{perf_hM2.fig}. It can be seen that for the longitudinal velocity estimation the performance is close for both observers with a slight advantage to the proposed approach due to the glitches seen in the estimations of the GRU-based observer \cite{srinivasan_end--end_2020}; for the lateral velocity estimation and the yaw rate estimation, the proposed approach is able to follow the reference iTraceRT sensor values more closely. None of the considered state of the art approaches is able to achieve low estimation errors for the three variables simultaneously.

%\begin{figure}[h!]
%    \centering
%    \includegraphics[width=\columnwidth]{81_A.pdf}
%    \caption{Lateral acceleration evolution along the near-limits dynamic maneuver.}
%    \label{highMan2.fig}
%\end{figure}

\begin{table}
\centering
\setlength\tabcolsep{5 pt}
    \caption{Mean absolute error for the different observers calculated for the near-limits maneuvers. The errors of the proposed approach are higher than the previous driving maneuver but are the lowest among the state of the art observers. Cf. \Cref{res1.tab} for legend and color codes.}
    \begin{tabular}{|c|c|c|c|c|c|}
      \hline
      State & DBM \cite{van_aalst_adaptive_2018} & 4WM \cite{katriniok_adaptive_2016}  & KN \cite{escoriza_data-driven_2021}  & GRU \cite{srinivasan_end--end_2020}  & Ours\\
      \hline
      $V_x \SI{}{(m/s)}$ & \cellcolor{red} 0.48 & 0.13 & 0.10 & \cellcolor{orange} 0.091 &\cellcolor{green!70} \textbf{0.079} \\
      \hline
      $V_y \SI{}{(m/s)}$ & \cellcolor{red} 0.18 & 0.10 & 0.10 & \cellcolor{orange} 0.068 & \cellcolor{green!70} \textbf{0.065} \\
      \hline
      $\dot\psi \SI{}{(mrad/s)}$ &\cellcolor{orange} 15.8 & 17.0 &\cellcolor{red} 18.5 & 16.1 & \cellcolor{green!70} \textbf{9.2} \\
      \hline
    \end{tabular}
\label{res4.tab}
\end{table}

In summary, the presented approach adapts to the high dynamics of the maneuver and accurately estimates $V_x$, $V_y$ and $\dot\psi$ of the vehicle for normal and high dynamic driving. Our approach outperforms the considered state of the art approaches by delivering the lowest errors for \emph{all three} estimated state variables in all cases.

\section{Conclusion}\label{conclusion.sec}
In this work we presented a novel LSTM-based observer architecture to estimate the longitudinal and lateral velocities and the yaw rate of the vehicle. We validated our method on real vehicle data thoroughly collected to reflect low dynamic and high dynamic scenarios. The observer takes into consideration its previous estimations and the measurements for the previous 50 time steps. The presented approach was tested and compared to state of the art model-based and learning-based observers while considering different driving conditions: normal city driving and near-limits driving. After thorough analysis of the results we proved that the proposed method is able to accurately estimate the three variables of interest while adapting to the harshness of the maneuver. The presented observer clearly outperforms state of the art observers.

Future work would explore the creation of a more diverse training set, acceleration wise; the effects of training on a specific acceleration range will be analyzed.  

\bibliographystyle{ieeetr}
\bibliography{mylib}

\end{document}